\pdfoutput=1

\documentclass[11pt]{article}

\usepackage[final]{acl}

\usepackage{times}
\usepackage{latexsym}

\usepackage[T1]{fontenc}

\usepackage[utf8]{inputenc}

\usepackage{microtype}

\usepackage{inconsolata}

\usepackage{graphicx}

\usepackage{amsmath}
\usepackage{amsfonts}
\usepackage{booktabs}
\usepackage{multirow} %
\usepackage{tabularx}
\usepackage{graphicx}
\usepackage{xargs}
\usepackage{makecell}
\usepackage{xcolor}
\newcommand{\blue}[1]{\textcolor{blue}{#1}}

\usepackage{algorithm,algorithmicx}
\usepackage[noend]{algpseudocode}

\algnewcommand\algorithmicinput{\textbf{Input:}}
\algnewcommand\Input{\item[\algorithmicinput]}
\algnewcommand\algorithmicoutput{\textbf{Output:}}
\algnewcommand\Output{\item[\algorithmicoutput]}
\usepackage{float}
\usepackage{bm}
\usepackage{bbm}
\usepackage{comment}
\usepackage{svg}
\usepackage{mathtools}
\usepackage[inline]{enumitem}
\usepackage{subcaption}
\usepackage{pgfplots}
\usepackage{amsmath}
\usepackage{sansmath}  
\usepackage{xcolor}  
\usepgfplotslibrary{groupplots}
\pgfplotsset{compat=newest}
\newcommand{\interalia}[1]{\citep[\emph{inter alia}]{#1}}

\title{Mitigating Shortcut Learning with InterpoLated Learning}

\author{
Michalis Korakakis$^{1,3}$ \ \ \ \ \ Andreas Vlachos$^{1}$ \ \ \ \ \ Adrian Weller$^{2,3}$  \\
 $^1$Department of Computer Science and Technology, University of Cambridge\\
 $^2$Department of Engineering, University of Cambridge\\
 $^3$The Alan Turing Institute \\
  {\texttt{\{mk2008,av308,aw665\}@cam.ac.uk}} \\
}

\begin{document}
\maketitle

\begin{abstract}
Empirical risk minimization~(ERM) incentivizes models to exploit shortcuts, i.e., spurious correlations between input attributes and labels that are prevalent in the majority of the training data but unrelated to the task at hand. This reliance hinders generalization on minority examples, where such correlations do not hold. Existing shortcut mitigation approaches are model-specific, difficult to tune, computationally expensive, and fail to improve learned representations. To address these issues, we propose InterpoLated Learning~(InterpoLL) which interpolates the representations of majority examples to include features from intra-class minority examples with shortcut-mitigating patterns. This weakens shortcut influence, enabling models to acquire features predictive across both minority and majority examples. Experimental results on multiple natural language understanding tasks demonstrate that InterpoLL improves minority generalization over both ERM and state-of-the-art shortcut mitigation methods, without compromising accuracy on majority examples. Notably, these gains persist across encoder, encoder-decoder, and decoder-only architectures, demonstrating the method's broad applicability.
\end{abstract}

\begin{figure*}[h]
    \centering
    \includegraphics[width=\textwidth]{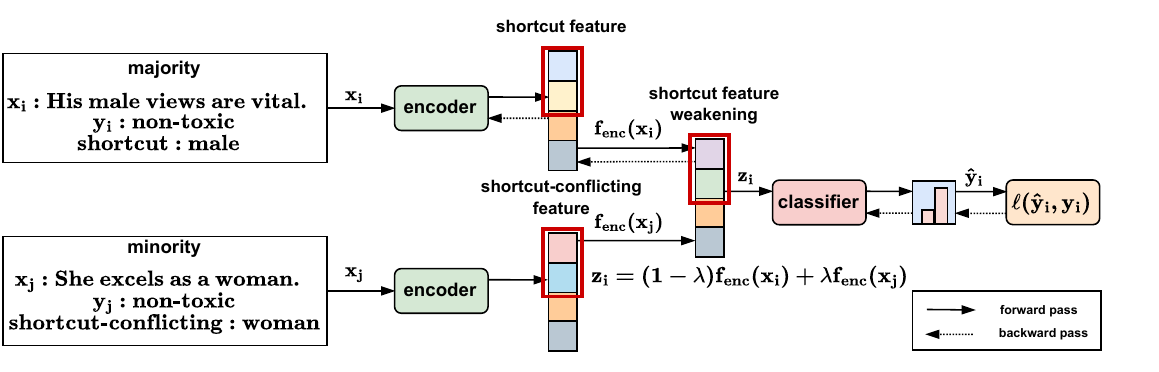}
    \caption{Illustration of InterpoLated Learning~(InterpoLL) for a majority example $x_{i}$. The term ``male'' in $x_{i}$ is over-represented in the ``non-toxic'' class, creating a shortcut that models exploit. Conversely, the minority example $x_{j}$, also labeled ``non-toxic,'' contains the under-represented term ``female,'' which counters this shortcut. To mitigate shortcut reliance, during the forward pass for $x_{i}$, we interpolate between the encoder representations of $x_{i}$ and $x_{j}$ as $z_{i} = (1-\lambda) f_{\text{enc}}(x_{i}) + \lambda f_{\text{enc}}(x_{j})$. This interpolation reduces the influence of shortcut features while preserving the label of $x_{i}$, as $x_j$ shares the same ground-truth label. The interpolated representation $z_{i}$ is then used in the backward pass for $x_{i}$ to compute the loss and update model parameters.}    
\label{fig:interpoll}
\end{figure*}

\section{Introduction}
Models trained via empirical risk minimization~(ERM) are prone to using shortcuts~\citep{DBLP:journals/natmi/GeirhosJMZBBW20}, i.e., superficial patterns between input features and labels that occur in the majority of training data due to idiosyncrasies in dataset collection and annotation, yet are irrelevant to the task at hand. For example, in MNLI~\citep{williams-etal-2018-broad}, most ``entailment'' examples exhibit high word overlap between the premise and the hypothesis~\citep{mccoy-etal-2019-right}. As a result, ERM-trained models achieve high in-distribution~(ID) accuracy by relying on this shortcut, but fail on minority examples where the heuristic breaks down, such as instances where high word overlap corresponds to ``contradiction.'' Reliance on shortcuts impairs out-of-distribution~(OOD) generalization, particularly under distribution shifts, where minority examples are underrepresented in training but more prevalent at test time~\citep{DBLP:conf/icml/YangZKG23}. Shortcut learning is widespread across NLP tasks, including text classification~\citep{park-etal-2018-reducing, DBLP:conf/aies/DixonLSTV18, DBLP:conf/www/BorkanDSTV19} and question answering~\citep{jia-liang-2017-adversarial, DBLP:conf/aaai/ShinodaSA23, mikula-etal-2024-think}. This highlights a fundamental issue: ERM-trained models often solve the dataset rather than the underlying task.

Typical shortcut mitigation methods include synthetically augmenting minority examples~\interalia{zhao-etal-2018-gender, min-etal-2020-syntactic, DBLP:conf/cikm/LeeWKLPJ21, wu-etal-2022-generating}, and learning an example weight distribution to up-weight minority examples~\citep{utama-etal-2020-towards, DBLP:conf/iclr/Sanh0BR21, DBLP:conf/icml/LiuHCRKSLF21, korakakis-vlachos-2023-improving}. However, these techniques primarily improve the classification layer operating on top of model representations~\citep{DBLP:conf/nips/IzmailovKGW22}. In fact, they do not learn representations different from those of ERM and can even reinforce shortcut features within the model~\citep{mendelson-belinkov-2021-debiasing}. Additionally, many shortcut mitigation methods rely on auxiliary models, increasing computational overhead and hyper-parameter tuning complexity~\citep{DBLP:conf/icml/PezeshkiBIBVL24}. The introduction of auxiliaries can further destabilize training  when their learning dynamics differ from those of the primary learner model~\citep{amirkhani-pilehvar-2021-dont-discard}.

To address these limitations, we propose InterpoLated Learning~(InterpoLL), an interpolation-based technique that mitigates shortcut learning and improves generalization on minority examples. In contrast to existing methods that rely on group annotations, which are impractical to obtain in real-world settings, InterpoLL operates without requiring prior knowledge of minority or majority group labels. InterpoLL interpolates the representations of majority examples with those of intra-class minority examples that exhibit shortcut-mitigating features~(Figure~\ref{fig:interpoll}). This discourages the model from relying on shortcuts for prediction, and shifts learning toward features that are predictive across all examples.

We evaluate InterpoLL on six natural language understanding datasets. Results show that InterpoLL achieves substantial improvements in minority generalization over ERM and state-of-the-art methods. Notably, it outperforms approaches that rely on prior knowledge of minority examples in the training data. InterpoLL also enhances domain generalization without access to domain-specific data. These improvements are consistent across encoder, encoder-decoder, and decoder-only architectures, as well as different model scales.

Detailed analyses demonstrate that, compared to existing shortcut mitigation methods, InterpoLL: (1) integrates with multiple techniques for identifying minority and majority examples; (2) reduces the presence of shortcut features in model representations; (3) demonstrates greater robustness to noise in the training data; (4) balances training dynamics between minority and majority examples; and (5) maintains a runtime comparable to ERM.

\section{Method}\label{sec:interpoll}

\begin{algorithm*}[th!]
\caption{InterpoLated Learning~(InterpoLL)}
\label{algo:interpoll}
\begin{algorithmic}[1]
    \Input Training dataset $\mathcal{D}$, auxiliary model $f_{{\phi}}$, learner model $f_\theta = \{f_{\text{enc}}, f_{\text{cls}}\}$
    
    \State{$g_\text{min}, g_\text{maj}= \textsc{InferMinMaj}(f_{{\phi}}, \mathcal{D})$}\Comment{classify examples into minority and majority}
    \State{Sample $B_1 \sim \mathcal{D}$} 
    \State{$B_2 = \emptyset$}    
    \For{$(x_i, y_i) \in B_1$}
    \If{$(x_i, y_i) \in g_\text{maj}$}
    \State{Sample $(x_j, y_j) \sim g_\text{min}$ s.t. $y_{i} = y_{j}$}\label{algo:sample}\Comment{sample intra-class minority example}
    \State{Sample $\lambda \sim \mathrm{Uniform}(0, 0.5)$}\label{algo:lambda}\Comment{sample interpolation ratio}
    \State{$z_{i} \gets(1-\lambda) f_{\text{enc}}(x_{i}) + \lambda f_{\text{enc}}(x_{j})$}\label{algo:interpolate}\Comment{add minority features to majority example}
    \Else
    \State{$z_{i} \gets f_{\text{enc}}(x_{i})$}\label{algo:minority}\Comment{use $f_{\text{enc}}$ for minority examples}
    \EndIf      
    \State{$B_2 \gets B_2 \cup (x_i, z_{i}, y_i)$}
    \EndFor
    \State{Update $\theta$ using $B_2$}
\end{algorithmic}
\end{algorithm*}

\subsection{Preliminaries}
\paragraph{Problem Setting} 
We consider a supervised learning task where $x \in \mathcal{X}$ is the input and $y \in \mathcal{Y}$ is the ground-truth label. Following~\citet{DBLP:conf/icml/SagawaRKL20}, we further assume that the training dataset contains distinct groups of examples within some classes. Some of these groups are well-represented and strongly associated with labels, e.g., high word overlap and ``entailment'' in natural language inference~(NLI) datasets~\citep{mccoy-etal-2019-right}, while others are not, e.g., negation words in the hypothesis and ``entailment''~\citep{gururangan-etal-2018-annotation}. We refer to the examples belonging to the well-represented groups associated with a particular class as majority examples $g_{\text{maj}}$ of said class, and the rest as minority $g_{\text{min}}$.\footnote{Note that some examples can be majority for a particular class, and other examples exhibiting the same patterns can be minority for another, e.g., in NLI, examples containing negation words in the hypothesis are majority for the ``contradiction'' class, but minority for the ``entailment'' class.}

We want to train a model $f_\theta: \mathcal{X} \rightarrow \mathcal{Y}$ parameterized by $\theta$. The model $f_\theta$ consists of an encoder $f_\text{enc}: \mathcal{X} \rightarrow \mathcal{Z}$ mapping the input $x_{i}$ to the representation $z_{i}$, and a classifier $f_\text{cls}: \mathcal{Z} \rightarrow \mathcal{Y}$ which takes the representation $z_{i}$ as input and maps it to a softmax probability over $\mathcal{Y}$. The model $f_\theta$ is typically optimized to minimize the average training loss through ERM. Given a loss function $\ell(f_\theta(x_i), y_i)$, e.g., cross-entropy loss, ERM minimizes the following objective: 

\begin{align}
    J_{\text{ERM}}(\theta) = \frac{1}{n} \sum \limits_{i=1}^n \ell(f_\theta(x_i), y_i)\,.
    \label{eq:erm}
\end{align}

Training with ERM causes the model to emphasize shortcut features prevalent in majority examples~\citep{DBLP:journals/corr/abs-2308-12553}. However, at test time, minority examples with features contradicting the shortcuts may become more prevalent~\citep{DBLP:conf/icml/KohSMXZBHYPGLDS21}. As a result, an ERM-trained model generalizes poorly.

\paragraph{Inferring Minority/Majority Examples}
We assume no prior knowledge about the minority and majority examples in the training dataset. To infer them, we use an under-parameterized ERM-trained auxiliary model $f_{{\phi}}$, as under-parameterization exacerbates model reliance on shortcut features within majority examples~\citep{ DBLP:conf/iclr/Sanh0BR21, DBLP:conf/icml/LiuHCRKSLF21}. Formally, we categorize examples in the training dataset $\mathcal{D}$ as minority if they are misclassified by $f_{{\phi}}$, i.e., $(x_{i}, y_{i}) \in g_{\text{min}} \text{ iff } f_{\phi}(x_{i}) \neq y_{i}$, and as majority if they are correctly classified by $f_{{\phi}}$, i.e., $(x_{i}, y_{i}) \in g_{\text{maj}} \text{ iff } f_{\phi}(x_{i}) = y_{i}$.

\subsection{InterpoLated Learning}
Here, we address the limitation of the ERM training objective, which fails to produce models that perform consistently across all examples in a dataset, by introducing InterpoLated Learning~(InterpoLL). Our goal is to enable the learner model $f_{\theta}$ to focus on features that are predictive across all examples. To achieve this, we interpolate the representations of majority examples with those from minority examples within the same class, whose features contradict the shortcuts. This process weakens the influence of shortcuts, preventing the model from relying on them for predictions. The overall procedure of InterpoLL is detailed in Algorithm~\ref{algo:interpoll}.

Concretely, for each majority example $(x_{i}, y_{i}) \in g_{\text{maj}}$ in the current mini-batch, we randomly select a minority example $(x_{j}, y_{j})$ from the identified set of minority examples $g_{\text{min}}$, ensuring that $y_{i} = y_{j}$~(line~\ref{algo:sample}). Then we interpolate their representations~(line~\ref{algo:interpolate}):

\begin{align}
    z_{i}= (1-\lambda) f_{\text{enc}}(x_{i}) + \lambda f_{\text{enc}}(x_{j})\,,
\label{eq:interpoll}
\end{align}

\noindent where the interpolation ratio $\lambda$ is sampled from $\mathrm{Uniform}(0, 0.5)$. This process diminishes the shortcut features within $x_{i}$ by incorporating shortcut-mitigating representations from $x_{j}$. Sampling $\lambda$ from $\mathrm{Uniform}(0, 0.5)$ ensures that the representation of $x_{i}$ is minimally altered, allowing the model to fit majority examples without exploiting shortcuts~(line~\ref{algo:lambda}). The label of the interpolated majority example $x_{i}$ remains unchanged, as both $x_{i}$ and $x_{j}$ belong to the same class.

\begin{table*}[h!]
    \centering
    \setlength{\tabcolsep}{2.25pt}
    \small
    \begin{tabular}{l|ccc|ccc|ccc|ccc}
    \toprule
    \multirow{2}{*}{\textbf{Method}}  & \multicolumn{3}{c|}{\textbf{MNLI}} & \multicolumn{3}{c|}{\textbf{FEVER}} & \multicolumn{3}{c|}{\textbf{QQP}} & \multicolumn{3}{c}{\textbf{Avg}} \\
    & ID & OOD & Stress & ID & OOD & Stress & ID & OOD & Stress & ID & OOD & Stress \\
    \midrule
    ERM & 84.9{\scriptsize$\pm$0.3} & 62.4{\scriptsize$\pm$1.3} & 62.9{\scriptsize$\pm$0.8} & 88.4{\scriptsize$\pm$0.4} & 55.9{\scriptsize$\pm$1.6} & 62.8{\scriptsize$\pm$2.5} & 90.2{\scriptsize$\pm$0.5} & 33.8{\scriptsize$\pm$0.6} & 63.5{\scriptsize$\pm$1.8} & 87.8 & 50.7 & 63.1 \\
    \midrule
    \multicolumn{13}{c}{\textbf{Prior Minority/Majority Example Information Required}} \\
    \midrule    
    GroupDRO & 84.3{\scriptsize$\pm$0.4} & 72.5{\scriptsize$\pm$0.6} & 63.1{\scriptsize$\pm$1.5} & 87.5{\scriptsize$\pm$0.4} & 64.1{\scriptsize$\pm$0.5} & 64.8{\scriptsize$\pm$1.7} & 89.5{\scriptsize$\pm$0.3} & 52.9{\scriptsize$\pm$0.8} & 66.2{\scriptsize$\pm$1.2} & 87.1 & 63.2 & 64.7 \\
    PoE & 83.8{\scriptsize$\pm$0.7} & 69.7{\scriptsize$\pm$1.5} & 61.9{\scriptsize$\pm$0.8} & 86.5{\scriptsize$\pm$0.6} & 61.9{\scriptsize$\pm$1.8} & 63.2{\scriptsize$\pm$2.1} & 89.1{\scriptsize$\pm$0.6} & 51.3{\scriptsize$\pm$1.3} & 64.1{\scriptsize$\pm$1.5} & 86.5 & 61.0 & 63.1 \\
    Conf-reg  & 84.7{\scriptsize$\pm$0.6} & 71.9{\scriptsize$\pm$1.8} & 62.5{\scriptsize$\pm$1.3} & 87.1{\scriptsize$\pm$0.4} & 62.4{\scriptsize$\pm$1.9} & 63.9{\scriptsize$\pm$0.8} & 89.8{\scriptsize$\pm$0.4} & 52.7{\scriptsize$\pm$1.5} & 65.3{\scriptsize$\pm$1.6} & 87.2 & 62.3 & 63.9 \\
    JTT & 84.1{\scriptsize$\pm$0.4} & 71.8{\scriptsize$\pm$0.5} & 62.8{\scriptsize$\pm$0.9} & 87.1{\scriptsize$\pm$0.3} & 62.3{\scriptsize$\pm$1.1} & 64.2{\scriptsize$\pm$0.9} & 89.3{\scriptsize$\pm$0.5} & 52.5{\scriptsize$\pm$0.6} & 65.5{\scriptsize$\pm$1.4} & 86.8 & 62.2 & 64.2 \\
    DFR & 84.4{\scriptsize$\pm$0.5} & 72.3{\scriptsize$\pm$0.9} & 63.1{\scriptsize$\pm$1.1} & 87.2{\scriptsize$\pm$0.3} & 62.7{\scriptsize$\pm$0.7} & 64.6{\scriptsize$\pm$1.2} & 89.4{\scriptsize$\pm$0.4} & 53.5{\scriptsize$\pm$1.3} & 65.9{\scriptsize$\pm$0.6} & 87.0 & 62.8 & 64.5 \\
    \midrule
    \multicolumn{13}{c}{\textbf{No Prior Minority/Majority Example Information Required}} \\
    \midrule
    Minimax & 83.5{\scriptsize$\pm$0.4} & 72.4{\scriptsize$\pm$0.6} & 62.6{\scriptsize$\pm$1.4} & 85.3{\scriptsize$\pm$0.5} & 62.6{\scriptsize$\pm$0.5} & 64.5{\scriptsize$\pm$0.8} & 87.9{\scriptsize$\pm$0.7} & 53.8{\scriptsize$\pm$1.6} & 65.8{\scriptsize$\pm$1.4} & 85.6 & 62.9 & 64.3 \\
    Weak-learn & 83.8{\scriptsize$\pm$0.8} & 71.8{\scriptsize$\pm$1.3} & 62.3{\scriptsize$\pm$0.6} & 85.6{\scriptsize$\pm$0.3} & 62.1{\scriptsize$\pm$1.1} & 64.1{\scriptsize$\pm$1.3} & 88.3{\scriptsize$\pm$0.3} & 53.2{\scriptsize$\pm$1.1} & 65.7{\scriptsize$\pm$1.7} & 85.9 & 62.4 & 64.0 \\
     CNC & 83.1{\scriptsize$\pm$1.4} & 72.4{\scriptsize$\pm$1.1} & 62.7{\scriptsize$\pm$2.1} & 84.7{\scriptsize$\pm$0.9} & 62.8{\scriptsize$\pm$1.5} & 63.7{\scriptsize$\pm$2.5} & 88.9{\scriptsize$\pm$0.8} & 53.5{\scriptsize$\pm$1.2} & 64.6{\scriptsize$\pm$2.1} & 85.6 & 62.9 & 63.7 \\
     RNF & 83.6{\scriptsize$\pm$0.8} & 69.9{\scriptsize$\pm$0.6} & 62.2{\scriptsize$\pm$1.7} & 85.1{\scriptsize$\pm$0.8} & 61.4{\scriptsize$\pm$1.4} & 64.2{\scriptsize$\pm$2.4} & 87.6{\scriptsize$\pm$0.5} & 52.2{\scriptsize$\pm$0.7} & 64.2{\scriptsize$\pm$1.8} & 85.4 & 61.2 & 63.5 \\
    \textbf{InterpoLL} & 84.6{\scriptsize$\pm$0.5} & \underline{\textbf{75.6}}{\scriptsize$\pm$0.7} & \underline{\textbf{66.5}}{\scriptsize$\pm$0.8} & 87.8{\scriptsize$\pm$0.6} & \underline{\textbf{68.7}}{\scriptsize$\pm$0.8} & \underline{\textbf{69.8}}{\scriptsize$\pm$1.1} & 89.8{\scriptsize$\pm$0.5} & \underline{\textbf{56.9}}{\scriptsize$\pm$1.0} & \underline{\textbf{70.5}}{\scriptsize$\pm$0.6} & 87.4 & \textbf{67.1}{\blue{$\uparrow$3.9}} & \textbf{68.9}{\blue{$\uparrow$4.2}} \\
    \bottomrule
    \end{tabular}
    \caption{Accuracies on the MNLI, FEVER, and QQP datasets. We report results on in-distribution development sets~(ID), out-of-distribution test sets~(OOD), and stress test sets~(Stress). All numbers are averaged over five runs (with different random seeds), and standard deviations are reported. The best-performing results are bolded, while underlining indicates statistically significant improvements over the ERM-trained baseline (t-test, $p < 0.05$). Values in blue denote improvements over the next best result.}
    \label{tab:main_shortcuts_nli}
\end{table*}

Then, we train the learner $f_{\theta}$ as in ERM~(Equation~\ref{eq:erm}) with the following modification: during the forward pass for a majority example $(x_i, y_i) \in g_\text{maj}$, instead of using the representation provided by $f_{enc}$, we use $z_i$ as obtained in Equation~\ref{eq:interpoll}. The backward pass, however, proceeds as normal, i.e., $z_i$ is used to compute the loss and gradients that update all the parameters of the learner model $f_{\theta}$. Conversely, for minority examples, we do not alter their representations via Equation~\ref{eq:interpoll}; thus, both the forward and backward passes use the representations provided by $f_{\text{enc}}$~(line~\ref{algo:minority}).

\paragraph{Intuition} InterpoLL can be viewed as a form of implicit data augmentation, as it repeatedly adds representations from intra-class minority examples to majority examples. This process increases the prevalence of minority example features, leading to a more uniform distribution of examples.

\section{Experimental Setup}\label{sec:minority}
\paragraph{Datasets}
We conduct experiments on six English datasets covering NLI and text classification. For NLI, we use MNLI~\citep{williams-etal-2018-broad}, FEVER~\citep{thorne-etal-2018-fever}, and QQP~\citep{Chen2017QuoraQP}, and evaluate minority generalization using OOD and stress test sets~\citep{naik-etal-2018-stress}. For text classification, we use FDCL18~\citep{DBLP:conf/icwsm/FountaDCLBSVSK18}, CivilComments-WILDS, and Amazon-WILDS~\citep{DBLP:conf/icml/KohSMXZBHYPGLDS21}, and evaluate minority generalization based on minority-group accuracy on the test set.

\paragraph{Comparisons}
We compare InterpoLL against shortcut mitigation methods that require prior knowledge of minority and majority examples: \textbf{GroupDRO}~\citep{DBLP:journals/corr/abs-1911-08731}, \textbf{PoE}~\citep{karimi-mahabadi-etal-2020-end}, \textbf{Conf-reg}~\citep{utama-etal-2020-mind},  \textbf{JTT}~\citep{DBLP:conf/icml/LiuHCRKSLF21}, and \textbf{DFR}~\citep{DBLP:conf/iclr/KirichenkoIW23}. We also evaluate InterpoLL against shortcut mitigation methods that do not rely on such knowledge: \textbf{Weak-learn}~\citep{DBLP:conf/iclr/Sanh0BR21}, \textbf{Minimax}~\citep{korakakis-vlachos-2023-improving}, \textbf{CNC}~\citep{DBLP:conf/icml/ZhangSZFR22}, and \textbf{RNF}~\citep{DBLP:conf/nips/DuMWTAH21}.

\paragraph{Implementation Details}
We use BERT-base~\citep{devlin-etal-2019-bert} from HuggingFace~\citep{DBLP:conf/emnlp/WolfDSCDMCRLFDS20} as the learner model, and TinyBERT~\citep{jiao-etal-2020-tinybert} as the auxiliary model in InterpoLL for our main experiments. We extract sentence embeddings from the CLS token of the final layer to perform interpolations. Each experiment is conducted five times with different random seeds, and results are reported as mean accuracy with standard deviations.

\begin{table*}[h!]
    \centering
    \setlength{\tabcolsep}{4.5pt}
    \small
    \begin{tabular}{l|cc|cc|c|cc}
    \toprule
    \multirow{2}{*}{\textbf{Method}} & \multicolumn{2}{c|}{\textbf{FDCL18}} & \multicolumn{2}{c|}{\textbf{CivilComments-WILDS}} & \textbf{Amazon-WILDS} & \multicolumn{2}{c}{\textbf{Avg}}  \\
    & Average & Minority & Average & Minority & 10th Percentile & Average & Minority \\
    \midrule
    ERM & 81.3\scriptsize{$\pm$0.2} & 35.6\scriptsize{$\pm$2.5} & 85.2\scriptsize{$\pm$0.3} & 63.5\scriptsize{$\pm$0.9} & 53.8\scriptsize{$\pm$0.8} & 83.2 & 49.6 \\
    \midrule
    \multicolumn{8}{c}{\textbf{Prior Minority/Majority Example Information Required}} \\
    \midrule
    GroupDRO & 76.2\scriptsize{$\pm$1.5}  & 57.3\scriptsize{$\pm$3.4}  & 82.1\scriptsize{$\pm$0.4} & 69.5\scriptsize{$\pm$0.9} & 53.3\scriptsize{$\pm$0.0} & 79.2 & 63.4 \\ 
    PoE      & 69.5\scriptsize{$\pm$2.3} & 53.1\scriptsize{$\pm$2.7} & 83.8\scriptsize{$\pm$0.7} & 69.6\scriptsize{$\pm$0.8} & 52.9\scriptsize{$\pm$0.8}  & 76.6 & 61.4 \\
    Conf-reg & 76.3\scriptsize{$\pm$1.2} & 55.9\scriptsize{$\pm$1.9} & 84.2\scriptsize{$\pm$0.5} & 67.2\scriptsize{$\pm$1.2} & 53.4\scriptsize{$\pm$0.8}  & 80.2 & 61.6 \\
    JTT & 76.1\scriptsize{$\pm$1.4} & 55.6\scriptsize{$\pm$1.7} & 83.9\scriptsize{$\pm$0.4} & 69.9\scriptsize{$\pm$0.8} & 52.9\scriptsize{$\pm$0.8} & 80.0 & 62.8 \\
    DFR & 76.3\scriptsize{$\pm$1.2} & 56.9\scriptsize{$\pm$1.4} & 83.7\scriptsize{$\pm$0.6} & 69.8\scriptsize{$\pm$0.7} & 53.4\scriptsize{$\pm$0.8} & 80.0  & 63.4 \\
    \midrule
    \multicolumn{8}{c}{\textbf{No Prior Minority/Majority Example Information Required}} \\
    \midrule
    Minimax & 75.9\scriptsize{$\pm$1.1} & 56.8\scriptsize{$\pm$1.8} & 83.2\scriptsize{$\pm$0.4} & 68.4\scriptsize{$\pm$0.8} & 52.9\scriptsize{$\pm$0.0}  & 79.6 & 62.6 \\
    Weak-learn & 71.6\scriptsize{$\pm$1.3} & 56.5\scriptsize{$\pm$2.0} & 83.1\scriptsize{$\pm$0.5} & 67.9\scriptsize{$\pm$0.7} & 52.4\scriptsize{$\pm$0.0} & 77.4 & 62.2 \\
    CNC & 75.6\scriptsize{$\pm$0.7} & 56.7\scriptsize{$\pm$1.3} & 83.5\scriptsize{$\pm$0.9} & 70.1\scriptsize{$\pm$1.4} & 53.4\scriptsize{$\pm$0.0} & 79.6 & 63.4 \\
    RNF & 72.3\scriptsize{$\pm$0.8} & 54.9\scriptsize{$\pm$1.6} & 83.8\scriptsize{$\pm$0.4} & 67.1\scriptsize{$\pm$0.9} & 52.9\scriptsize{$\pm$0.8} & 78.0 & 61.0 \\
    \textbf{InterpoLL} & 78.8\scriptsize{$\pm$0.6} & \underline{\textbf{61.2}}\scriptsize{$\pm$0.7} & 84.7\scriptsize{$\pm$0.7} & \underline{\textbf{73.9}}\scriptsize{$\pm$0.3} & \underline{\textbf{55.2}}\scriptsize{$\pm$0.0}  & 81.8 & {\textbf{67.6}}{\blue{$\uparrow$4.2}} \\
    \bottomrule
    \end{tabular}
    \caption{Model performance on the FDCL18, CivilComments-WILDS, and Amazon-WILDS datasets. For FDCL18 and CivilComments-WILDS, we report average and minority test accuracy, and for Amazon-WILDS, we report 10th Percentile accuracy~\citep{DBLP:conf/icml/KohSMXZBHYPGLDS21}. All numbers are averaged over five runs (with different random seeds), and standard deviations are reported. The best-performing results are bolded, while underlining indicates statistically significant improvements over the ERM-trained baseline (t-test, $p < 0.05$). The value in blue indicates improvement over the next best result. Amazon-WILDS is excluded from the calculations in the last two columns.}
    \label{tab:main_shortcuts_text} 
\end{table*}

\section{Main Results}
\paragraph{Natural Language Inference} Based on the results in Table~\ref{tab:main_shortcuts_nli}, we observe that although all shortcut mitigation methods improve minority generalization compared to ERM, no single previously proposed method consistently outperforms the others. Among methods that require prior knowledge of minority and majority examples, GroupDRO achieves the best OOD and stress test accuracy. Methods that do not rely on such prior information generally perform worse, except for InterpoLL, which achieves the highest OOD and stress test accuracy overall. Compared to GroupDRO, the next best method, InterpoLL improves OOD accuracy by 3.9\% and stress test accuracy by 4.2\%.

\paragraph{Text Classification} Table~\ref{tab:main_shortcuts_text} presents the average and minority test set accuracies for FDCL18 and CivilComments-WILDS, and the 10th percentile accuracy for Amazon-WILDS, following~\citet{DBLP:conf/icml/KohSMXZBHYPGLDS21}. Similar to the results from NLI, the effectiveness of previous approaches varies across these datasets. However, InterpoLL consistently outperforms all methods, achieving the highest minority accuracy, 4.2\% higher than the next best approach, GroupDRO.

\section{Domain Generalization}\label{sec:ood}
\paragraph{Experimental Setup} We focus on domain generalization, where a model is applied to out-of-domain settings without access to any additional domain data. We hypothesize that by reducing the model's reliance on shortcut features for predictions, InterpoLL facilitates the acquisition of task-relevant features, thereby enhancing domain generalization. To this end, we conduct experiments using GLUE-X~\citep{yang-etal-2023-glue}. We restrict our comparison to methods that do not rely on prior assumptions about shortcuts, as GLUE-X does not provide information on minority or majority examples. Detailed dataset statistics are provided in Table~\ref{tab:out-of-domain-stats} in the Appendix.

\begin{table}[!h]
\centering
\small
\setlength{\tabcolsep}{0.8pt}
\begin{tabular}{lcccccc|c}
\toprule
\textbf{Method} & \textbf{SST-2} & \textbf{MNLI} & \textbf{QNLI} & \textbf{RTE} & \textbf{MRPC} & \textbf{QQP}  & \textbf{Avg} \\
\midrule
ERM            & 87.2  & 68.7  & 77.1  & 57.9  & 58.4  & 66.5  & 69.3 \\
Minimax        & 88.1  & 70.2  & 77.3  & 57.7  & 58.8  & 67.5  & 69.9 \\
Weak-learn     & 87.5  & 68.9  & 77.5  & 57.8  & 58.6  & 66.7  & 69.5 \\
RNF            & 87.8  & 69.8  & 77.4  & 56.2  & 58.4  & 66.8  & 69.4 \\
\textbf{InterpoLL} & \textbf{90.1} & \textbf{71.5} & \textbf{81.2} & \textbf{60.6} & \textbf{60.9} & \textbf{69.8} & \textbf{72.4}{\blue{$\uparrow$2.5}} \\
\bottomrule
\end{tabular}
\caption{Domain generalization results on GLUE-X. The value in blue indicates an improvement over the next best result.}
\label{tab:out-of-domain}
\end{table}

\paragraph{Results} Table~\ref{tab:out-of-domain} shows domain generalization results. We observe that Minimax, Weak-learner, and RNF offer only minor improvements over ERM, with average gains of 0.6\%, 0.2\%, and 0.1\%, respectively. In contrast, InterpoLL achieves average improvements of 3.1\% over ERM, 2.5\% over Minimax, 2.9\% over Weak-learner, and 3.0\% over RNF.

\section{Analysis}\label{sec:interpoll_analysis}

\begin{table}[!h]
    \centering
    \scriptsize
    \setlength{\tabcolsep}{2pt} 
    \renewcommand{\arraystretch}{0.85} 
    \begin{tabular}{p{1.5cm} p{1.2cm} c c c c c c c}
        \toprule 
        \textbf{Model} & \textbf{Method} & \textbf{HANS} & \textbf{PAWS} & \textbf{Sym.} & \textbf{RTE} & \textbf{MRPC} & \textbf{Avg} \\
        \midrule
        BERT-l & \makecell{ERM \\ InterpoLL} & \makecell{73.8 \\ \textbf{78.2}} & \makecell{42.9 \\ \textbf{59.4}} & \makecell{65.2 \\ \textbf{69.1}} & \makecell{64.9 \\ \textbf{68.6}} & \makecell{61.7 \\ \textbf{63.4}} & \makecell{61.7 \\ \textbf{67.7}} \\
        \midrule
        RoBERTa-l & \makecell{ERM \\ InterpoLL} & \makecell{74.7 \\ \textbf{80.1}} & \makecell{45.4 \\ \textbf{60.5}} & \makecell{68.1 \\ \textbf{74.5}} & \makecell{74.4 \\ \textbf{77.3}} & \makecell{64.6 \\ \textbf{66.9}} & \makecell{65.4 \\ \textbf{71.9}} \\
        \midrule
        XLNet-l & \makecell{ERM \\ InterpoLL} & \makecell{75.3 \\ \textbf{81.6}} & \makecell{47.6 \\ \textbf{59.7}} & \makecell{69.4 \\ \textbf{78.2}} & \makecell{75.1 \\ \textbf{78.4}} & \makecell{64.3 \\ \textbf{67.5}} & \makecell{66.4 \\ \textbf{73.1}} \\
        \midrule
        ELECTRA-l & \makecell{ERM \\ InterpoLL} & \makecell{76.2 \\ \textbf{81.8}} & \makecell{49.3 \\ \textbf{61.6}} & \makecell{70.5 \\ \textbf{78.4}} & \makecell{76.3 \\ \textbf{77.9}} & \makecell{68.4 \\ \textbf{70.3}} & \makecell{68.1 \\ \textbf{74.0}} \\
        \midrule
        T5-l & \makecell{ERM \\ InterpoLL} & \makecell{77.8 \\ \textbf{82.9}} & \makecell{50.1 \\ \textbf{63.9}} & \makecell{71.8 \\ \textbf{78.9}} & \makecell{79.1 \\ \textbf{81.5}} & \makecell{70.8 \\ \textbf{72.7}} & \makecell{69.9 \\ \textbf{76.0}} \\
        \midrule
        T5-3B & \makecell{ERM \\ InterpoLL} & \makecell{78.9 \\ \textbf{84.5}} & \makecell{51.2 \\ \textbf{65.1}} & \makecell{72.4 \\ \textbf{79.7}} & \makecell{79.9 \\ \textbf{82.8}} & \makecell{71.6 \\ \textbf{74.2}} & \makecell{70.8 \\ \textbf{77.3}} \\
        \midrule
        GPT2-m & \makecell{ERM \\ InterpoLL} & \makecell{61.2 \\ \textbf{67.8}} & \makecell{44.9 \\ \textbf{54.3}} & \makecell{64.7 \\ \textbf{69.2}} & \makecell{65.8 \\ \textbf{68.7}} & \makecell{56.1 \\ \textbf{59.5}} & \makecell{58.5 \\ \textbf{63.9}} \\
        \midrule
        GPT2-l & \makecell{ERM \\ InterpoLL} & \makecell{70.8 \\ \textbf{77.4}} & \makecell{47.5 \\ \textbf{59.6}} & \makecell{66.8 \\ \textbf{72.1}} & \makecell{74.9 \\ \textbf{77.8}} & \makecell{61.4 \\ \textbf{63.3}} & \makecell{64.3 \\ \textbf{70.0}} \\
        \bottomrule   
    \end{tabular}
    \caption{Performance for large models trained with ERM and InterpoLL. The experimental setup for MNLI, FEVER, and QQP follows Section~\ref{sec:minority}, while the setup for RTE and MRPC follows Section~\ref{sec:ood}. The best results are bolded.}
    \label{tab:large_models}
\end{table}

\paragraph{Comparison Across Model Sizes and Architectures} We investigate whether the generalization improvements achieved by InterpoLL extend to larger models and different architectures. Specifically, we evaluate its performance on large-scale encoder models, including BERT-large~\citep{devlin-etal-2019-bert}~($\approx$340M params), RoBERTa-large~\citep{DBLP:journals/corr/abs-1907-11692}~($\approx$355M params), ELECTRA-large~\citep{DBLP:conf/iclr/ClarkLLM20}~($\approx$340M params), and XLNet-large~\citep{DBLP:conf/nips/YangDYCSL19}~($\approx$340M params). We also evaluate encoder-decoder models, such as T5-large~\citep{DBLP:journals/jmlr/RaffelSRLNMZLL20}~($\approx$770M params) and T5-3B~\citep{DBLP:journals/jmlr/RaffelSRLNMZLL20}~($\approx$3B params), as well as decoder-only models, including GPT2-medium~\citep{Radford2019LanguageMA}~($\approx$350M params) and GPT2-large~\citep{Radford2019LanguageMA}~($\approx$750M params). For T5 models, interpolations are applied in the final layers of both the encoder and decoder, while for GPT2 models, interpolations are applied in the final layer of the decoder. The results in Table~\ref{tab:large_models} show that InterpoLL improves average OOD performance by 6.0 points for BERT-large, 6.5 points for RoBERTa-large, 6.7 points for XLNet-large, 5.9 points for ELECTRA-large, 6.1 points for T5-large, 6.5 points for T5-3B, 5.4 points for GPT2-medium, and 5.7 points for GPT2-large.

\begin{figure}[!t]
\centering
\begin{tikzpicture}
\begin{axis}[
    width=0.4\textwidth,
    height=0.36\textwidth,
    ybar=0pt,
    bar width=10pt,
    enlarge x limits=0.15,
    ylabel style={font=\small, align=center},
    tick label style={font=\footnotesize},
    ytick={65, 70, 75, 80, 85, 90},
    ymin=65, ymax=90,
    grid=major,
    grid style={dotted, gray!90},
    major tick length=0pt,
    tick style={black},
    symbolic x coords={erm_const, uni_const, beta_const, beta_const2},
    ylabel={\textbf{Accuracy (\%)}},
    xtick=data,
    xticklabels={$\mathrm{U(0, .5)}$, $\mathrm{U(0, 1)}$, $\mathrm{B(.5, .5)}$, $\mathrm{B(2, 2)}$},
        legend style={font=\footnotesize, at={(0.5,-0.2)}, anchor=north, legend columns=-1},
    legend cell align={left},
    xticklabel style={font=\footnotesize, yshift=0.5ex}
]
\addplot+[bar shift=-5pt, fill=blue!70, draw=black, line width=0.8pt] coordinates {(erm_const, 84.5) (uni_const, 77.9) (beta_const, 80.3) (beta_const2, 83.4)};
\addplot+[bar shift=5pt, fill=orange!70, draw=black, line width=0.8pt] coordinates {(erm_const, 75.6) (uni_const, 76.2) (beta_const, 72.2) (beta_const2, 75.9)};
\legend{\textcolor{blue!70}{ID}, \textcolor{orange!70}{OOD}}
\end{axis}
\end{tikzpicture}
\caption{In-distribution~(ID) accuracy on MNLI and out-of-distribution~(OOD) generalization on HANS for different interpolation ratios $\lambda$ used in InterpoLL: uniform distributions $\mathrm{U(0, 0.5)}$ and $\mathrm{U(0, 1)}$; beta distributions $\mathrm{B(0.5, 0.5)}$ and $\mathrm{B(2, 2)}$.}
\label{fig:interp_ratio}
\end{figure}
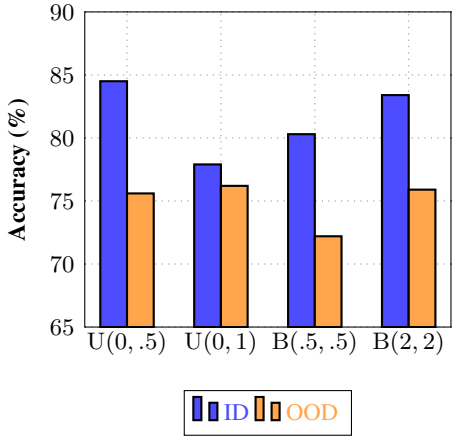

\paragraph{Impact of Interpolation Ratio} We investigate the impact of the interpolation ratio $\lambda$ on InterpoLL when it is sampled from four different \begin{enumerate*}[label=(\arabic*), before=\unskip{ distributions: }, itemjoin={{, }}, itemjoin*={{, and }}]
\item $\mathrm{Uniform}(0, 0.5)$
\item $\mathrm{Uniform}(0, 1)$
\item U-shaped $\mathrm{Beta}(0.5, 0.5)$
\item bell-shaped $\mathrm{Beta}(2, 2)$.
\end{enumerate*}
Figure~\ref{fig:interp_ratio} presents results on MNLI. We observe that sampling $\lambda$ from $\mathrm{Uniform}(0, 0.5)$ achieves better performance by sufficiently weakening the shortcuts, while still enabling the model to learn majority features. Other distributions cause a drastic drop in accuracy either by incorporating a higher proportion of minority features into majority examples, impairing the model's ability to fit them, or by failing to sufficiently weaken the shortcuts.

\begin{table}[!h]
        \centering
        \small
        \begin{tabular}{l|cc|cc}
        \toprule
        \multirow{2}{*}{\textbf{Method}}  & \multicolumn{2}{c|}{\textbf{MNLI}} & \multicolumn{2}{c}{\textbf{CivilComments}} \\
        & ID & HANS & Average & Minority  \\
        \midrule
        ERM & 84.9 & 62.4 & 85.2 & 63.5 \\
        no auxiliary & 84.5 & 75.3 & 84.9 & 73.5 \\
        BERT-tiny & 84.6 & \textbf{75.6} & 84.7 & \textbf{73.9} \\
        BERT-base & 84.1 & 75.2 & 84.5 & 73.6 \\
        BERT-large & 84.5 & 74.8 & 84.1 & 73.2 \\
        under-trained & 84.3 & 75.1 & 84.6 & 73.8 \\
        regularized & 84.4 & 74.9 & 84.4& 73.7 \\
        \bottomrule
        \end{tabular}        
        \caption{Accuracies on MNLI and CivilComments-WILDS using different auxiliary models for InterpoLL.}
        \label{tab:auxiliary_size} 
\end{table}

\paragraph{Impact of Auxiliary} To evaluate how auxiliary model choice affects InterpoLL's effectiveness, we experiment with BERT models of varying sizes. We also evaluate two additional variants: a BERT-base model trained with ERM for 3 epochs (under-trained)~\citep{utama-etal-2020-towards}, and a BERT-base model trained with weight decay set to 1 (regularized)~\citep{DBLP:conf/icml/LiuHCRKSLF21}. Lastly, we assess a variant of InterpoLL that operates without an auxiliary model, where the learner itself identifies minority and majority examples (no auxiliary)~\citep{korakakis-vlachos-2023-improving}. In this setup, the learner is trained for 2 epochs, and its misclassifications are used as a proxy for identifying minority and majority examples. Results from Table~\ref{tab:auxiliary_size} show that all auxiliary variants achieve comparable generalization. Notably, the no-auxiliary InterpoLL variant still outperforms the other shortcut mitigation methods on HANS and CivilComments-WILDS, as shown in Tables~\ref{tab:main_shortcuts_nli} and~\ref{tab:main_shortcuts_text}.

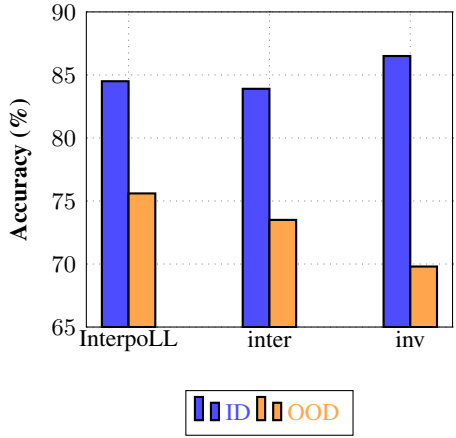
\begin{figure}[!t]
\centering
\begin{tikzpicture}
\begin{axis}[
    width=0.4\textwidth,
    height=0.36\textwidth,
    ybar=0pt,
    bar width=10pt,
    enlarge x limits=0.15,
    ylabel style={font=\small, align=center},
    tick label style={font=\footnotesize},
    ytick={65, 70, 75, 80, 85, 90},
    ymin=65, ymax=90,
    grid=major,
    grid style={dotted, gray!90},
    major tick length=0pt,
    tick style={black},
    ylabel={\textbf{Accuracy (\%)}},
    symbolic x coords={erm_const, uni_const, inter_const},
    xtick=data,
    xticklabels={InterpoLL, inter, inv},
    xticklabel style={font=\footnotesize, yshift=0.5ex},
    legend style={font=\footnotesize, at={(0.5,-0.2)}, anchor=north, legend columns=-1},
    legend cell align={left}
]
\addplot+[bar shift=-5pt, fill=blue!70, draw=black, line width=0.8pt] coordinates {(erm_const, 84.5)  (uni_const, 83.9) (inter_const, 86.5)};
\addplot+[bar shift=5pt, fill=orange!70, draw=black, line width=0.8pt] coordinates {(erm_const, 75.6) (uni_const, 73.5) (inter_const, 69.8)};
\legend{\textcolor{blue!70}{ID}, \textcolor{orange!70}{OOD}}
\end{axis}
\end{tikzpicture}
\caption{In-distribution~(ID) accuracy on MNLI and out-of-distribution~(OOD) generalization on HANS for two InterpoLL variants: inter, which performs inter-class interpolation between minority and majority examples; and inv, an inverse variant that weakens minority representations by interpolating them with those of majority examples.}
\label{fig:interpoll_variants}
\end{figure}

\paragraph{InterpoLL Variants} To better understand the effects of different components of InterpoLL, we conduct experiments with two \begin{enumerate*}[label=(\arabic*), before=\unskip{ variants: }, itemjoin={{, }}]
\item {inter-class InterpoLL~(inter)} interpolates the representations of majority examples with those of inter-class minority examples, in contrast to standard InterpoLL, which involves intra-class interpolations
\item {inverse InterpoLL~(inv)} reverses the interpolation direction, i.e., interpolating the representations of minority examples with those of majority examples, thus weakening the features of the former.
\end{enumerate*} Figure~\ref{fig:interpoll_variants} presents results for models trained on MNLI and evaluated on HANS. We find that intra-class interpolations outperform inter-class ones, as the latter may introduce irrelevant features into majority examples. Similarly, reversing the interpolation direction improves accuracy on majority examples but reduces it on minority examples.

\begin{table}[h!]
\centering
\setlength{\tabcolsep}{3pt}
\small
\begin{tabular}{l|c|c|c}
\toprule
\textbf{Method} & \textbf{Overlap$\downarrow$} & \textbf{Subsequence$\downarrow$} & \textbf{Negation$\downarrow$} \\ 
\midrule
ERM       & 2.9{\scriptsize$\pm$0.3}  & 2.5{\scriptsize$\pm$0.3}  & 2.7{\scriptsize$\pm$0.3}  \\ 
\midrule
GroupDRO & 3.1{\scriptsize$\pm$0.2} & 2.9{\scriptsize$\pm$0.5} & 2.9{\scriptsize$\pm$0.3}  \\
PoE      & 3.9{\scriptsize$\pm$0.2} & 4.1{\scriptsize$\pm$0.4} & 3.0{\scriptsize$\pm$0.3}  \\
Conf-reg & 4.5{\scriptsize$\pm$0.3} & 4.2{\scriptsize$\pm$0.3} & 3.2{\scriptsize$\pm$0.6}  \\
JTT      & 3.9{\scriptsize$\pm$0.2} & 4.1{\scriptsize$\pm$0.4} & 3.0{\scriptsize$\pm$0.3}  \\
DFR      & 3.0{\scriptsize$\pm$0.4} & 2.5{\scriptsize$\pm$0.5} & 2.8{\scriptsize$\pm$0.5}\\
Minimax  & 3.5{\scriptsize$\pm$0.2} & 3.2{\scriptsize$\pm$0.2} & 3.2{\scriptsize$\pm$0.3}  \\ 
Weak-learn & 3.4{\scriptsize$\pm$0.3} & 3.3{\scriptsize$\pm$0.5} & 3.2{\scriptsize$\pm$0.5} \\
CNC      & 2.8{\scriptsize$\pm$0.6} & 2.6{\scriptsize$\pm$0.4} & 2.6{\scriptsize$\pm$0.5}\\
RNF      & 3.3{\scriptsize$\pm$0.5} & 3.8{\scriptsize$\pm$0.5} & 3.2{\scriptsize$\pm$0.3}\\
\midrule
\textbf{InterpoLL} & \textbf{2.7}{\scriptsize$\pm$0.3} & \textbf{2.5}{\scriptsize$\pm$0.3} & \textbf{2.5}{\scriptsize$\pm$0.2}  \\
\bottomrule
\end{tabular}
\caption{Probing results for Overlap, Subsequence, and Negation shortcut categories on MNLI. Higher values indicate greater extractability of shortcut features from model representations.}
\label{tab:mnli_bias_extractability}
\end{table}

\paragraph{Shortcut Extractability}
We evaluate the extractability of shortcut features from model representations using minimum description length probing~\citep{voita-titov-2020-information, mendelson-belinkov-2021-debiasing, DBLP:conf/aaai/LyuLYRRZYR23}. We report compression, where higher values suggest increased shortcut extractability from the model representations. Table~\ref{tab:mnli_bias_extractability} shows the probing results on MNLI. We observe that prior methods increase the extractability of shortcuts from model representations, as indicated by high compression values. InterpoLL exhibits the lowest compression values, while achieving the highest minority generalization performance~(Table~\ref{tab:main_shortcuts_nli}). Table~\ref{tab:mnli_bias_extractability_full_results} in the Appendix presents detailed results.

\begin{table}[]
        \centering
        \small
        \begin{tabular}{l|cc|cc}
        \toprule
        \multirow{2}{*}{\textbf{Method}}  & \multicolumn{2}{c|}{\textbf{MNLI}} & \multicolumn{2}{c}{\textbf{CivilComments}} \\
        & ID & HANS & Average & Minority  \\
        \midrule
        ERM & 41.8 & 38.1 & 44.3 & 40.1 \\
        GroupDRO & 35.9 & 34.7 & 39.5 & 37.7 \\
        RNF & 38.3 & 37.6 & 37.6 & 35.2 \\
        JTT & 36.9 & 33.4 & 38.9 & 36.4 \\
        Minimax & 35.1 & 34.3 & 37.9 & 37.3 \\
        InterpoLL & \textbf{64.8} & \textbf{63.9} & \textbf{62.8} & \textbf{61.3} \\
        \bottomrule
        \end{tabular}        
        \caption{Model performance on MNLI and CivilComments-WILDS with 5\% synthetic label noise in the training data.}
        \label{tab:noise} 
\end{table}

\paragraph{Robustness to Noise} To assess InterpoLL's robustness under noisy conditions, we introduce 5\% synthetic label noise into the training data and evaluate performance on MNLI and CivilComments-WILDS. The results in Table~\ref{tab:noise} indicate that InterpoLL exhibits substantially greater resilience to noise compared to ERM and other shortcut mitigation methods.

\begin{table}[!h]
    \centering
    \small
    \setlength{\tabcolsep}{4pt}
    \renewcommand{\arraystretch}{1.1}
    \begin{tabular}{l c}
        \toprule
        \textbf{Method} & \textbf{Time (hr)} \\
        \midrule
        ERM & 4 \\
        DFR & 4 \\
        PoE & 6 \\
        Weak-learn & 6 \\
        Conf-reg & 7 \\
        RNF & 5 \\
        JTT & 8 \\
        CNC & 39 \\
        Minimax & 5 \\
        InterpoLL & 5 \\
        \bottomrule
    \end{tabular}
    \caption{Training runtime (in hours) on MNLI.}
    \label{tab:runtime}
\end{table}

\paragraph{Training Runtime} We assess the efficiency of shortcut mitigation methods by measuring training time. As shown in Table~\ref{tab:runtime}, InterpoLL introduces only a minor computational overhead compared to ERM, while most other methods approximately double ERM's training time. Notably, InterpoLL is substantially faster than CNC, which also targets the reduction of shortcut features in model representations.

\begin{figure*}[!t]
\centering
\begin{tikzpicture}
\begin{groupplot}[
    group style={
        group size=2 by 1,
        horizontal sep=1.5cm,
    },
    width=0.42\textwidth,
    height=0.3\textwidth,
    xlabel={\textbf{Epochs}},
    ylabel={\textbf{Accuracy (\%)}},
    xlabel style={font=\small},
    ylabel style={font=\small},
    tick label style={font=\small},
    title style={font=\small\bfseries},
    ymin=40, ymax=100,
    xtick={0,2,4,6,8,10},
    ytick={0,20,40,60,80,100},
    grid=major,
    grid style={line width=.1pt, draw=gray!20},
    legend to name=commonlegend,
    legend style={font=\small, draw=black, fill=none},
]

\nextgroupplot[title={\textbf{MNLI}}]
ylabel={\textbf{Accuracy (\%)}},
\addplot[color=blue!80, mark=square*, line width=1pt] coordinates {(0, 80) (1, 85) (2, 84) (3, 83) (4, 83) (5, 84) (6, 84) (7, 84) (8, 85) (9, 84) (10, 83)};
\addlegendentry{Avg ERM}

\addplot[color=red, mark=triangle*, line width=1pt, dashed] coordinates {(0, 70) (1, 75) (2, 74) (3, 78) (4, 77) (5, 80) (6, 82) (7, 82) (8, 83) (9, 82) (10, 81)};
\addlegendentry{Avg InterpoLL}

\addplot[color=orange!80, mark=pentagon*, line width=1pt, solid] coordinates {(0, 49) (1, 62) (2, 62) (3, 65) (4, 65) (5, 67) (6, 63) (7, 64) (8, 64) (9, 64) (10, 65)};
\addlegendentry{Min ERM}

\addplot[color=purple, mark=diamond*, line width=1pt, dash dot] coordinates {(0, 63) (1, 69) (2, 70) (3, 73) (4, 74) (5, 76) (6, 75) (7, 77) (8, 76) (9, 77) (10, 78)};
\addlegendentry{Min InterpoLL}

\nextgroupplot[title={\textbf{CivilComments-WILDS}}]
ylabel={},
\addplot[color=blue!80, mark=square*, line width=1pt] coordinates {(0, 86) (1, 87) (2, 88) (3, 88) (4, 87) (5, 87) (6, 87) (7, 88) (8, 88) (9, 87) (10, 86)};
\addplot[color=red, mark=triangle*, line width=1pt, dashed] coordinates {(0, 74) (1, 76) (2, 79) (3, 77) (4, 80) (5, 83) (6, 83) (7, 83) (8, 84) (9, 82) (10, 82)};
\addplot[color=orange!80, mark=pentagon*, line width=1pt, solid] coordinates {(0, 44) (1, 49) (2, 59) (3, 61) (4, 60) (5, 58) (6, 62) (7, 61) (8, 61) (9, 60) (10, 61)};
\addplot[color=purple, mark=diamond*, line width=1pt, dash dot] coordinates {(0, 65) (1, 69) (2, 69) (3, 72) (4, 71) (5, 73) (6, 74) (7, 73) (8, 72) (9, 71) (10, 72)};

\end{groupplot}
\end{tikzpicture}

\begin{tikzpicture}
\begin{axis}[
    hide axis,
    xmin=0, xmax=1,
    ymin=0, ymax=1,
    legend columns=4,
    legend style={font=\small, draw=black, fill=white},
]
\addlegendimage{color=blue!80, mark=square*, line width=1pt}
\addlegendentry{Avg ERM}
\addlegendimage{color=red, mark=triangle*, line width=1pt, dashed}
\addlegendentry{Avg InterpoLL}
\addlegendimage{color=orange!80, mark=pentagon*, line width=1pt, solid}
\addlegendentry{Min ERM}
\addlegendimage{color=purple, mark=diamond*, line width=1pt, dash dot}
\addlegendentry{Min InterpoLL}
\end{axis}
\end{tikzpicture}
\ref{commonlegend}
\caption{Training set accuracy over epochs for minority and majority examples on MNLI and CivilComments-WILDS, using ERM and InterpoLL.}
\label{fig:dynamics}
\end{figure*}
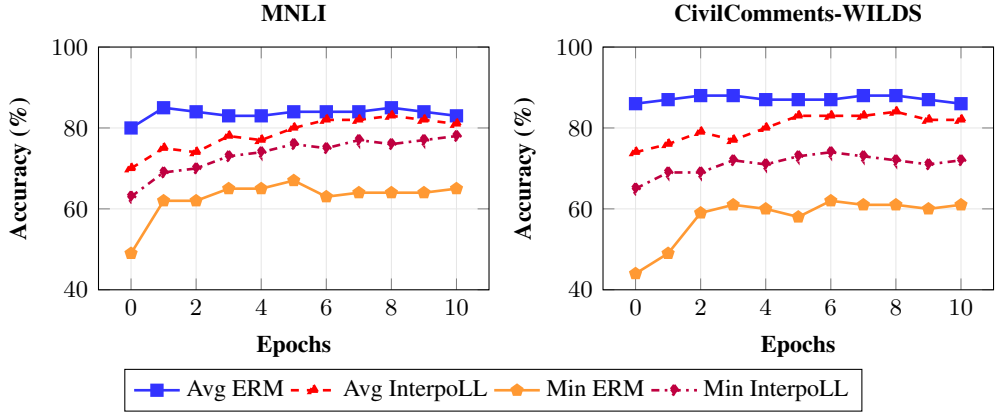

\begin{table*}[h!]
\centering
\small
\begin{tabular}{l|cccc}
\toprule
\textbf{Dataset} & \textbf{BERT-Tiny} & \textbf{BERT-Large} & \textbf{Under-trained} & \textbf{Regularised} \\ 
\hline
MNLI                & 95.1\% & 93.9\% & 94.6\% & 94.1\% \\ 
FEVER               & 97.1\% & 95.4\% & 96.3\% & 96.1\% \\ 
QQP                 & 94.9\% & 93.7\% & 93.5\% & 92.7\% \\ 
CivilComments-WILDS & 96.8\% & 94.8\% & 95.9\%  & 95.9\% \\ 
\bottomrule
\end{tabular}
\caption{Recall of minority examples across different datasets and auxiliary model variants.}
\label{tab:minority_recall}
\end{table*}

\paragraph{Model Training Dynamics} We analyze model training dynamics to assess InterpoLL's ability to balance training across minority and majority examples, thereby counteracting the tendency of ERM-trained models to focus on fitting shortcut features within majority examples first. Figure~\ref{fig:dynamics} shows the training set accuracy over epochs on the MNLI and CivilComments-WILDS datasets. We see that while an ERM-trained model eventually achieves high accuracy for minority examples, this only occurs towards the end of training. InterpoLL demonstrates more balanced training throughout, mitigating the model's initial preference for fitting majority examples.

\paragraph{Effectiveness of Minority Example Identification} To verify the reliability of different auxiliary models in inferring minority examples, we compute recall, defined as the proportion of ground-truth minority examples correctly identified. We conduct experiments on the MNLI, FEVER, QQP, and CivilComments-WILDS datasets. Table~\ref{tab:minority_recall} indicates that all auxiliary models achieve high recall.

\begin{table*}[!h]
\centering
\small
\setlength{\tabcolsep}{5pt}
\begin{tabular}{l@{\hskip 4pt}|cc|cc|c|c}
\toprule
\multirow{2}{*}{\textbf{Method}} & \multicolumn{2}{c|}{\textbf{MNLI}} & \multicolumn{2}{c|}{\textbf{CivilComments-WILDS}} & \textbf{Amazon-WILDS} & \textbf{GLUE-X} \\
& ID & HANS & Avg & Minority & 10th \% & \\ 
\midrule
ERM & 84.9$\pm$0.3 & 62.4$\pm$1.3 & 85.2$\pm$0.3 & 63.5$\pm$0.9 & 53.8$\pm$0.0 & 69.3 \\ 
\midrule
mixup & 84.7$\pm$0.3 & 69.3$\pm$1.4 & 84.9$\pm$0.2 & 66.9$\pm$1.1 & 52.4$\pm$0.0 & 69.4 \\ 
LISA & 84.5$\pm$0.5 & 72.2$\pm$0.8 & 84.4$\pm$0.1 & 70.7$\pm$0.4 & 53.9$\pm$0.0 & -- \\ 
\midrule
\textbf{InterpoLL (ours)} & 84.6$\pm$0.5 & \textbf{75.6}$\pm$0.7 & 84.7$\pm$0.7 & \textbf{73.9}$\pm$0.3 & \textbf{55.2}$\pm$0.0 & \textbf{72.4} \\ 
\bottomrule
\end{tabular}
\caption{Model performance on MNLI, CivilComments-WILDS, Amazon-WILDS, and GLUE-X~(averaged over five runs). InterpoLL and ERM results for MNLI and CivilComments-WILDS are repeated from Tables~\ref{tab:main_shortcuts_nli} and~\ref{tab:main_shortcuts_text}. LISA requires group annotations and is therefore not applicable to GLUE-X.}
\label{tab:interpolation_baselines}
\end{table*}

\paragraph{Weakening Shortcut Features vs Augmenting Minority Examples} We compare InterpoLL with mixup~\citep{DBLP:conf/iclr/ZhangCDL18} and LISA~\citep{DBLP:conf/icml/Yao0LZL0F22}, two interpolation-based data augmentation techniques. Mixup performs interpolations between random pairs of examples, whereas LISA interpolates examples that either have the same label but come from different example groups, or have different labels but the same example group. Conversely, InterpoLL targets the weakening of shortcut features prevalent in majority examples by interpolating their representations with those of intra-class minority examples exhibiting shortcut-mitigating features. To ensure that majority representations are not substantially altered, the interpolation ratio $\lambda$ is sampled from a $\mathrm{Uniform}(0, 0.5)$ distribution, rather than from $\mathrm{Uniform}(0, 1)$.

Table~\ref{tab:interpolation_baselines} presents the experimental results on the MNLI and CivilComments-WILDS datasets. On MNLI, InterpoLL achieves a 6.3\% improvement in terms of minority accuracy compared to mixup, and a 3.4\% improvement compared to LISA. Similarly, on CivilComments-WILDS, InterpoLL outperforms mixup and LISA by 7.0\% and 3.2\%, respectively. These results underscore InterpoLL's effectiveness in improving minority generalization over interpolation-based data augmentation methods.

\section{Related Work}

\paragraph{Mixup} InterpoLL is inspired by mixup~\citep{DBLP:conf/iclr/ZhangCDL18}, a data augmentation method that generates synthetic examples by interpolating random pairs of examples and their labels. Recent mixup extensions apply interpolations in representation space~\citep{DBLP:conf/icml/VermaLBNMLB19}, explore alternative interpolation strategies~\citep{DBLP:conf/aaai/GuoMZ19, DBLP:journals/tnn/MaiHCSS22, yin-etal-2021-batchmixup, DBLP:conf/uai/CollinsBLPSLW23}, and integrate mixup with regularization techniques~\citep{jeong-etal-2022-augmenting, kong-etal-2022-dropmix}. In NLP,~\citet{DBLP:journals/corr/abs-1905-08941} use mixup on word and sentence embeddings, whereas,~\citet{yoon-etal-2021-ssmix} propose a mixup variant that conducts interpolations on the input text. Several other works leverage mixup for model calibration~\citep{DBLP:journals/corr/abs-2102-11402, park-caragea-2022-calibration, park-caragea-2022-data}. \citet{zhang-etal-2020-seqmix} perform mixup to augment the unlabeled pool in active learning, while~\citet{korakakis-etal-2024-alvin} use it to identify informative and diverse unlabeled instances for annotation.

\paragraph{Shortcut Mitigation in NLP} Existing approaches for mitigating shortcuts can be categorized into two main groups, model- and data-centric~\citep{DBLP:journals/corr/abs-2208-11857}. Model-centric approaches typically assume that examples with shortcuts are learned more quickly than minority ones due to ERM training. To this end, a popular approach is ensemble an auxiliary model that relies on prior shortcut knowledge for predictions with a learner model during training, such that the latter is discouraged from using shortcuts~\citep{he-etal-2019-unlearn, clark-etal-2019-dont, stacey-etal-2020-avoiding, DBLP:conf/nips/XiongCPCML21, ghaddar-etal-2021-end}. More recently,~\citet{honda-etal-2024-eliminate} extended this setup in a post-hoc manner by combining the predictions of multiple models under the assumption that each model exploits different features for prediction. Other model-centric methods include contrastive learning~\citep{DBLP:conf/aaai/LyuLYRRZYR23}, knowledge distillation~\citep{du-etal-2021-towards}, forgettable examples~\citep{yaghoobzadeh-etal-2021-increasing}, adversarial training~\citep{belinkov-etal-2019-adversarial}, and pruning~\cite{meissner-etal-2022-debiasing}. Data-centric methods aim to reduce the presence of shortcuts by modifying the training data itself, typically through data augmentation~\citep{DBLP:conf/aaai/WangC21, stacey-rei-2024-distilling, cheng-amiri-2024-fairflow} or example filtering~\citep{DBLP:conf/icml/BrasSBZPSC20, swayamdipta-etal-2020-dataset}.

\section{Conclusion}
In this work, we introduced InterpoLL, an interpolation-based technique designed to mitigate shortcut learning and improve minority generalization without compromising accuracy on majority examples. The effectiveness of InterpoLL is demonstrated across various datasets, covering a range of natural language understanding tasks. Our analyses show that InterpoLL enhances domain generalization, remains effective across different model architectures and scales, and reduces shortcut features in model representations.

\section*{Limitations}
Similar to other shortcut mitigation methods, InterpoLL can result in reduced ID accuracy. Moreover, while our probing experiments demonstrate that InterpoLL reduces shortcut-feature extractability, our analysis does not examine  whether this suppression inadvertently impacts other aspects of the learned representations. Exploring such potential trade-offs is an important direction for future work. Finally, our experiments focus on specific English-language natural language understanding tasks, leaving InterpoLL's broader applicability to other tasks and languages unexplored.

\section*{Acknowledgments}
Michalis Korakakis is supported by the Cambridge Commonwealth, European and International Trust, the ESRC Doctoral Training Partnership, and the Alan Turing Institute. Andreas Vlachos is supported by the ERC grant AVeriTeC (GA 865958). Adrian Weller acknowledges support from a Turing AI Fellowship under grant EP/V025279/1.
\bibliography{anthology, custom}
\clearpage

\appendix

\section{Dataset Details}

\begin{table}[H]
\centering
\begin{tabular}{lll}
\toprule
\textbf{Task} & \textbf{ID} & \textbf{OOD} \\ 
\midrule
\multirow{4}{*}{\begin{tabular}[c]{@{}l@{}}Sentiment \\ Analysis\end{tabular}} 
 & \multirow{4}{*}{SST-2} & IMDB \\
 &  & Yelp \\
 &  & Amazon \\
 &  & Flipkart \\
\midrule
\multirow{3}{*}{\begin{tabular}[c]{@{}l@{}}Natural \\ Language \\ Inference\end{tabular}} 
 & \multirow{3}{*}{MNLI-m} & MNLI-mm \\
 &  & SNLI \\
 &  & SICK \\
\midrule
\begin{tabular}[c]{@{}l@{}}Question\\Answering \\NLI\end{tabular} 
 & QNLI & NewsQA \\
\midrule
\multirow{2}{*}{\begin{tabular}[c]{@{}l@{}}Textual \\ Entailment\end{tabular}} 
 & \multirow{2}{*}{RTE} & SciTail \\
 &  & HANS \\
\midrule
\multirow{4}{*}{Paraphrase} 
 & \multirow{2}{*}{MRPC} & QQP \\
 &  & Twitter \\
\cmidrule(lr){2-3} 
 & \multirow{2}{*}{QQP} & MRPC \\
 &  & Twitter \\
\bottomrule
\end{tabular}
\caption{Dataset statistics for out-of-domain generalization experiments on GLUE-X.}
\label{tab:out-of-domain-stats}
\end{table}

\section{Additional Analysis}

\begin{table}[!h]
\centering
\small
\setlength{\tabcolsep}{4pt}
\begin{tabular}{lcc}
\toprule
\textbf{Method} & \textbf{MNLI ID} & \textbf{HANS} \\
\midrule
ERM & 58.9 & 54.3 \\
InterpoLL & 65.3 & 71.9 \\
\bottomrule
\end{tabular}
\caption{Performance comparison on MNLI and HANS when the dominant shortcuts are removed from the training data.}
\label{tab:mnli_hans_hard}
\end{table}

\subsection{Hard-to-learn Shortcuts} 
To evaluate InterpoLL in settings where no known easy-to-learn shortcuts are present, we removed all MNLI training examples containing: (i) high word overlap between premise and hypothesis with the ``entailment'' label, and (ii) negation in the hypothesis with the ``contradiction'' label. As shown in Table~\ref{tab:mnli_hans_hard}, InterpoLL maintains its strong performance, outperforming ERM in both ID and OOD generalisation.

\begin{table}[!h]
\centering
\small
\setlength{\tabcolsep}{4pt}
\begin{tabular}{lcccc}
\toprule
\textbf{Method} & \textbf{p = 0.2} & \textbf{p = 0.4} & \textbf{p = 0.6} & \textbf{p = 0.8} \\
\midrule
ERM & 80.1 & 75.2 & 60.9 & 55.9 \\
InterpoLL & 80.3 & 75.1 & 73.5 & 72.9 \\
\bottomrule
\end{tabular}
\caption{Performance comparison between ERM and InterpoLL for different values of $p$ for the synthetic shortcut.}
\label{tab:erm_interpoll_p}
\end{table}

\subsection{Synthetic Shortcut} 
We modified the MNLI training data by appending a prefix in the hypothesis containing the ground-truth label with probability $p$, or a random label otherwise. The modified test set always includes a random label in the prefix. We hypothesize that when the shortcut prevalence is low ($p$ < 0.5), the shortcut becomes harder to exploit as it no longer forms a dominant pattern. The results on the modified MNLI test set in Table~\ref{tab:erm_interpoll_p} confirm that InterpoLL remains effective even when shortcuts are less likely to be learned early on.

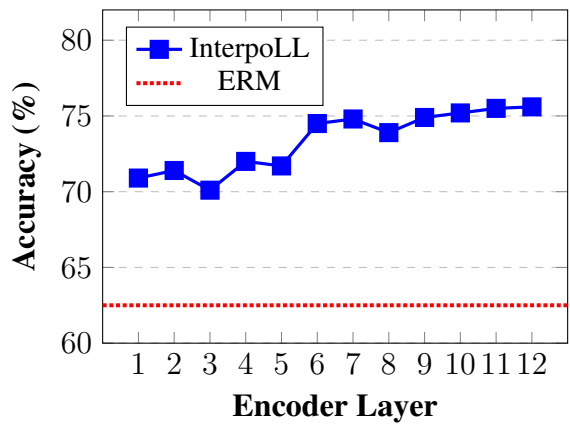
\begin{figure}[!h]
\centering
\begin{tikzpicture}
\begin{axis}[
    title={\bfseries },
    xlabel={\bfseries Encoder Layer},
    ylabel={\bfseries Accuracy (\%)},
    xmin=0, xmax=13, 
    ymin=60, ymax=80, 
    xtick={1,2,3,4,5,6,7,8,9,10,11,12}, 
    ytick={60,65,70,75,80}, 
    ymajorgrids=true, 
    grid style=dashed, 
    width=\columnwidth, 
    height=6cm, 
    tick label style={font=\bfseries\large}, 
    label style={font=\bfseries\large}, 
    title style={font=\bfseries\large}, 
    enlarge y limits={upper,value=0.1}, 
    cycle list name=color list, 
    legend style={
        at={(0.05,0.95)}, 
        anchor=north west
    }
]

\addplot[
    color=blue,
    mark=square*,
    line width=1.2pt,
    mark size=3pt
    ]
    coordinates {
    (1,70.9)(2,71.4)(3,70.1)(4,72.0)(5,71.7)(6,74.5)
    (7,74.8)(8,73.9)(9,74.9)(10,75.2)(11,75.5)(12,75.6)
    };
\addlegendentry{InterpoLL}

\addplot[
    color=red,
    mark=none,
    line width=1.5pt, 
    samples=2,
    domain=0:13, 
    style=densely dotted 
    ] {62.5};
\addlegendentry{ERM}

\end{axis}
\end{tikzpicture}
\caption{Model performance on HANS for different layers used in InterpoLL for the interpolations between minority and majority examples for a BERT-base model.}
\label{fig:interpolation_layers}
\end{figure}

\subsection{Impact of Interpolation Layer}~\label{appendix:enc_layder}
Figure~\ref{fig:interpolation_layers} illustrates the effect of using different encoder layers for conducting interpolations. Models are trained on MNLI and evaluated on HANS. We observe that the most effective layers for interpolations are situated in the upper regions of the model, where higher-level features are typically processed~\citep{jawahar-etal-2019-bert}. Notably, InterpoLL outperforms ERM in minority generalization, regardless of the encoder layer used for the interpolations.

\begin{table*}[]
\centering
\setlength{\tabcolsep}{1pt}
\small
\begin{tabular}{l|cc|cc|cc}
\toprule
& \multicolumn{2}{c|}{\textbf{Overlap}} & \multicolumn{2}{c|}{\textbf{Subsequence}} & \multicolumn{2}{c}{\textbf{Negation}}\\
\textbf{Method} & Compr. $\downarrow$ & Acc. $\downarrow$             & Compr. $\downarrow$               & Acc. $\downarrow$ & Compr. $\downarrow$               & Acc. $\downarrow$  \\ 
\midrule
ERM       & 2.9{\scriptsize$\pm$0.3}  & 86.9{\scriptsize$\pm$0.7}   & 2.5{\scriptsize$\pm$0.3}  & 87.8{\scriptsize$\pm$1.4} & 2.7{\scriptsize$\pm$0.3}  & 95.1{\scriptsize$\pm$0.5}  \\ 
\midrule
GroupDRO &   3.1{\scriptsize$\pm$0.2}   &  87.1{\scriptsize$\pm$0.6}     &    2.9{\scriptsize$\pm$0.5}   &  88.5{\scriptsize$\pm$1.3} &    2.9{\scriptsize$\pm$0.3}    & 96.1{\scriptsize$\pm$0.6}  \\
PoE  & 3.9{\scriptsize$\pm$0.2}       & 93.2{\scriptsize$\pm$1.4}         & 4.1{\scriptsize$\pm$0.4}        & 94.3{\scriptsize$\pm$2.1}  & 3.0{\scriptsize$\pm$0.3}        & 96.3{\scriptsize$\pm$0.7}  \\
Conf-reg    & 4.5{\scriptsize$\pm$0.3}         & 94.1{\scriptsize$\pm$0.5}         & 4.2{\scriptsize$\pm$0.3}         & 94.8{\scriptsize$\pm$1.9}  & 3.2{\scriptsize$\pm$0.6}        & 96.8{\scriptsize$\pm$0.5}   \\
JTT  & 3.9{\scriptsize$\pm$0.2}       & 93.2{\scriptsize$\pm$1.4}         & 4.1{\scriptsize$\pm$0.4}        & 94.3{\scriptsize$\pm$2.1}  & 3.0{\scriptsize$\pm$0.3}        & 96.3{\scriptsize$\pm$0.7}  \\
DFR  & 3.0{\scriptsize$\pm$0.4}       & 87.1{\scriptsize$\pm$0.7}         & 2.5{\scriptsize$\pm$0.5}        & 87.8{\scriptsize$\pm$1.5}  & 2.8{\scriptsize$\pm$0.5}        & 95.3{\scriptsize$\pm$0.4}  \\
Minimax  & 3.5{\scriptsize$\pm$0.2}    & 89.1{\scriptsize$\pm$0.8}    & 3.2{\scriptsize$\pm$0.2}   & 89.3{\scriptsize$\pm$1.9} & 3.2{\scriptsize$\pm$0.3}        & 96.5{\scriptsize$\pm$0.6} \\ 
Weak-learn & 3.4{\scriptsize$\pm$0.3}    & 88.9{\scriptsize$\pm$0.5}    & 3.3{\scriptsize$\pm$0.5}   & 89.9{\scriptsize$\pm$2.1} & 3.2{\scriptsize$\pm$0.5}       & 96.9{\scriptsize$\pm$0.7} \\
CNC & 2.8{\scriptsize$\pm$0.6}       & 86.8{\scriptsize$\pm$0.7}         & 2.6{\scriptsize$\pm$0.4}        & 87.7{\scriptsize$\pm$1.3}  & 2.6{\scriptsize$\pm$0.5}        & 94.9{\scriptsize$\pm$0.6}  \\
RNF & 3.3{\scriptsize$\pm$0.5}       & 87.3{\scriptsize$\pm$1.8}         & 3.8{\scriptsize$\pm$0.5}        & 92.4{\scriptsize$\pm$1.2}  & 3.2{\scriptsize$\pm$0.3}        & 96.5{\scriptsize$\pm$2.2}  \\
\midrule
\textbf{InterpoLL}  & \textbf{2.7}{\scriptsize$\pm$0.3}   & \textbf{86.7}{\scriptsize$\pm$0.4}  & \textbf{2.5}{\scriptsize$\pm$0.3}   & \textbf{87.6}{\scriptsize$\pm$1.5} & \textbf{2.5}{\scriptsize$\pm$0.2}  & \textbf{94.8}{\scriptsize$\pm$0.7}  \\
\bottomrule
\end{tabular}
\caption{Probing results for Overlap, Subsequence, and Negation shortcut categories on the MNLI dataset. Higher values in both compression~(Compr.) and accuracy~(Acc.) metrics indicate greater extractability of shortcut features from the model's representations.}
\label{tab:mnli_bias_extractability_full_results}
\end{table*}

\end{document}